\documentstyle[colap,fleqn,mylingmakros,short-long-ex,epsf,article-ex,array,newmakros]{article}

\title{\vspace{-0.5in}Yet Another Paper about Partial Verb Phrase Fronting in German}
\author{Stefan M\"uller\thanks{This paper is available via the WWW: {\tt http://
www.compling.hu-berlin.de/\~{}stefan/Pub/e\_pvp.html} Thanks to Frank Keller for comments on
earlier versions of this paper.
} \\
Humboldt University Berlin \\
Philosophische Fakult\"at II\\
Institut f\"ur deutsche Sprache und Linguistik \\
Lehrstuhl Computerlinguistik \\
J\"agerstra\ss{}e 10/11 \\
D-10099 Berlin \\
{\tt stefan@compling.hu-berlin.de}
}

\begin{document}

\mathindent0pt
\maketitle
\vspace{-0.5in}
\begin{abstract}

I describe a very simple HPSG analysis for partial verb phrase
fronting. I will argue that the presented account is more adequate
than others made during the past years
because it allows the description of constituents in fronted
positions with their modifier remaining in the non-fronted
part of the sentence.
 
A problem with ill-formed signs that are admitted by all HPSG accounts for
partial verb phrase fronting known so far will be explained and a solution
will be suggested that uses the difference between combinatoric relations
of signs and their representation in word order domains.

\end{abstract}

\section{Introduction}

During the last years, several different analyses for partial verb phrase fronting
have been proposed \cite{Pollard90,Nerbonne94,Baker94-Eng,HN94b}. The most promising account
so far has been the one of Hinrichs and Nakazawa. This account, however, suffers from
some drawbacks that will be discussed in section \ref{alternatives}. I will
present a rather simple account that uses the standard {\sc nonloc} mechanism HPSG \cite{ps2}
provides. In section \ref{licensing}, I will discuss a problem that arises for all
accounts of partial verb phrase fronting: underspecified {\sc comps} lists. By the means of
a new daughter (licensing daughter) in a schema for the introduction of nonlocal dependencies
this problem will be solved.

\section{The Phenomena}

In German, it is possible to front non-maximal verbal projections.\footnote{
	The examples (\ex{1}) and (\ex{3}) are taken from Hinrichs and Nakazawa (1994b).
}
\eenumsentence{
\label{bsp-erzaehlen-wird}
\item \longex{4}{1}
         {[Erz\"ahlen] & wird & er & seiner Tochter }{ ein M\"archen.}
	 {~tell        & will & he & his daughter   }{ a   fairy tail}
	 {`He will tell his daughter a fairy tale.'}
\item \longex{4}{2}
         {[Erz\"ahlen & m\"ussen] & wird & er }{ seiner Tochter & ein M\"archen.}
         {~tell       & must      & will & he }{ his daughter   & a fairy tale}
	 {`He will have to tell his daughter a \\fairy tale.'}
}
In a series of papers, Hinrichs and Nakazawa argued for a special rule schema
that combines the verbs of a so-called {\em verbal complex\/} before the arguments
of the involved verbs are combined with the verbal complex. 
Because the verbal complex is build before any nonverbal argument of a verb gets saturated,
it is possible to account for phenomena like {\it auxiliary flip\/}. As the verbal complex
is analyzed as a constituent, the fronting of {\em erz\"ahlen m\"ussen\/} in (\ex{0}b)
can be explained as well.
There is no problem with sentences like those in (\ex{0}) for the standard {\sc nonloc} mechanism.
{\em Erz\"ahlen m\"ussen\/} is a constituent in the non-fronted position in (\ex{1})
and the same holds if the verbal complex is fronted.
\enumsentence{
Er wird seiner Tochter ein M\"archen [erz\"ahlen m\"ussen].
}
There are, however, examples where a partly saturated verbal complex is
fronted.
\eenumsentence{
\label{bsp-maerchen}
\item {}[Seiner Tochter ein M\"archen erz\"ahlen] wird er.
\item {}[Ein M\"archen erz\"ahlen] wird er seiner Tochter.
\item {}[Ein M\"archen erz\"ahlen] wird er seiner Tochter m\"ussen.
\item {}[Seiner Tochter erz\"ahlen] wird er das M\"archen.
}
A verb with some of its arguments may appear in the {\it Vorfeld\/}
leaving other arguments in the {\it Mittelfeld\/}.

As (\ex{1}) shows, it is possible that a PP in the {\it Mittelfeld\/}
modifies a fronted verbal complex.
\enumsentence{
\longex{2}{3}
	{Den Kanzlerkandidaten    & ermorden }{ wollte & die Frau  & mit   diesem Messer.}
        {the chancellor.candidate & kill     }{ wanted & the woman & with  this   knife}
	{`The woman wanted to kill the candidate\\ with this knife.'}
}
Sentences like (\ex{1}a) are ungrammatical. It is not possible to front parts of the
verbal complex that would be located in the middle of the verbal complex in a verb
final sentence (\ex{1}b).
\eenumsentence{
\label{ex-muessen}
\item \longexnt{6}{2}
   { * & M\"ussen & wird & er & ihr }{ ein M\"archen & erz\"ahlen. }
   {   & must     & will & he & her }{ a   story     & tell        }
\item , weil er ihr ein M\"archen erz\"ahlen m\"ussen wird.
}

\section{The Analysis}

\subsection{Basic Assumptions}

In what follows, I assume a version of HPSG that deviates from standard HPSG
in that the surface string of a phrasal sign is not determined by a relation that
relates the {\sc phon} values of a sign to the {\sc phon} values 
of its daughters \cite[p. 169]{ps}. Instead I will follow
Reape's \shortcite{Reape94} approach. Reape assumes word order domains as
 an additional level of representation. In such a domain, all daughters of a head occur.
These domains differ from the daughter list in that the elements in a domain (signs)
correspond in their serialization to the surface order of the words in the string.
LP-constraints apply to elements of the order domain.
Another basic assumption of Reape is that constituents may be discontinuous.

As Hinrichs and Nakazawa (1994a) 
have shown, it is reasonable to assume in addition to the
head complement schema a schema that licenses the verbal complex. Hinrichs and Nakazawa
introduced the concept of argument attraction into the HPSG framework. If a verbal
complex is build two verbs are combined and the resulting sign inherits all arguments
from both verbs. In their paper, Hinrichs and Nakazawa treat verbal complements as
ordinary complements that are included in the {\sc comps} list of their heads. It has however
proven to be useful to distinguish the verbal complement from other complements 
\cite{Rentier94,Mueller95b-Eng}. The merits of this move will be discussed shortly.
For the purpose of representing the information about verbal complements, the feature
{\sc vcomp} is introduced.
Its value is
a {\it synsem\/}-object if the verb embeds another verb and {\it none\/} otherwise.
The entry in the stem lexicon for the future tense auxiliary {\em werden\/} ({\em will\/}) is shown in
(\ex{1}).
\begin{figure}[htbp]
\begin{equation}
\begin{array}{l}
\mbox{werden:} 
\\
\ms[cat]{
 head  & \ms[verb]{ subj & \ibox{1} \\
                   } \\[4mm]
 comps & \ibox{2} \\[1mm]
 vcomp & {\sc v[lex}{\rm +},{\it bse\/},{\sc subj}~\ibox{1},{\sc comps}~\ibox{2},\\
     ~ & ~~{\sc vcomp}~{\it none\/}{\sc ]} \\
}
\end{array}
\end{equation}
\end{figure}
From this stem the morphology component produces the finite form shown in (\ex{1}).
In German, almost any complement of a verb can be fronted, subjects as well as objects.
Therefore, for finite forms the subject is included into the {\sc comps} list, from where
extraction is possible. For nonfinite forms the subject does not appear on {\sc comps}
but stays in the {\sc subj} list.\footnote{see \cite{Kiss93} for details}
\begin{figure}[htbp]
\begin{equation}
\begin{array}{l}
\mbox{wird:}
\\
\ms[cat]{
 head   & \ms[verb]{ vform & fin \\
                     subj & \liste{} \\
                   } \\[5mm]
 comps & \ibox{1} $\oplus$ \ibox{2}\\[1mm] 
 vcomp  & {\sc V[{\sc lex}{\rm +\/},{\it bse\/},subj~\ibox{1}, comps~\ibox{2},} \\
     ~  & {\sc~~vcomp~{\it none}]} \\[1mm]
}
\end{array}
\end{equation}
\end{figure}
Schema \ref{schema-vk} licenses verb cluster structures.\footnote{
	I will not go into the details of the domain formation in verb cluster structures.
	For details see \cite{Mueller95-Eng}.
}
\begin{figure}[htbp]
\begin{samepage}
\begin{schema}[Verb Cluster Schema]
\label{schema-vk}
$
\\
\\
\ms[phrasal-sign]{
 synsem & \ms{ loc$|$cat & \ms{ vcomp & none \\
                            } \\[1mm]
               lex   & {\rm +} \\
             } \\[1mm]
 dtrs & \ms[head-cluster-structure]{
           head-dtr & \ms{ \ldots$|$vcomp & \ibox{1} \\[1mm]
                           dom & \ibox{2} \\
                         } \\[1mm]
           cluster-dtr & \ms{ synsem & \ibox{1} \\[1mm]
                              dom    & \ibox{3} \\
                            } \\[1mm]
           comp-dtrs   & \liste{} \\
         } \\[15mm]
 dom & \ibox{2} $\bigcirc$ \ibox{3} \\
}
$
\end{schema}
\end{samepage}
\end{figure}
A head is combined with its verbal complement (\ibox{1}). The resulting sign is a verbal 
complex or a part of a verbal complex. It is marked {\sc lex}+ because it can in turn be embedded.
\enumsentence{
\longex{4}{3}
      {, weil    & er & ihm & ein M\"archen }{ [[erz\"ahlen & lassen] & hat].}
      {~~because & he & him & a fairy tale  }{ ~~tell       & let     & has}
      { `because he has let him tell the story.'}
}

\subsection{The {\sc lex} Feature}

The {\sc lex} feature in the entry for {\em werden\/} ensures that a matrix verb is combined
with its verbal complement before the verbal complement is saturated by one of its complements.
It is therefore possible to avoid multiple structures in the {\it Mittelfeld\/}.
\eenumsentence{
\item Er wird seiner Tochter ein M\"archen [erz\"ahlen m\"ussen].
\item Er wird seiner Tochter [[ein M\"archen erz\"ahlen] m\"ussen]].
\item Er wird [[seiner Tochter ein M\"archen erz\"ahlen] m\"ussen]].
}
But exactly those constituents that have to be avoided in the {\it Mittelfeld\/}
are needed in the {\it Vorfeld\/}. Very complicate mechanisms have been introduced
to cope with this problem without a lot of spurious ambiguities \cite{Nerbonne94,HN94b}.
I will suggest a solution to the problem that is very simple: If 
it is the case that an embedded verb or verbal complex
has to be {\sc lex}+ when verb and complement are combined locally and if it is the case
that this does not hold if a nonlocal dependency is involved than the simplest solution
is to view {\sc lex} not as a local feature. If one assumes that {\sc lex} lives under the
path {\sc synsem} instead of {\sc synsem$|$loc} than the problem turns into a non-issue.\footnote{
	Detmar Meurers independently found the same solution.
}

\begin{figure*}[htbp]
\epsfxsize=0.9\textwidth
\centerline{\mbox{\epsffile{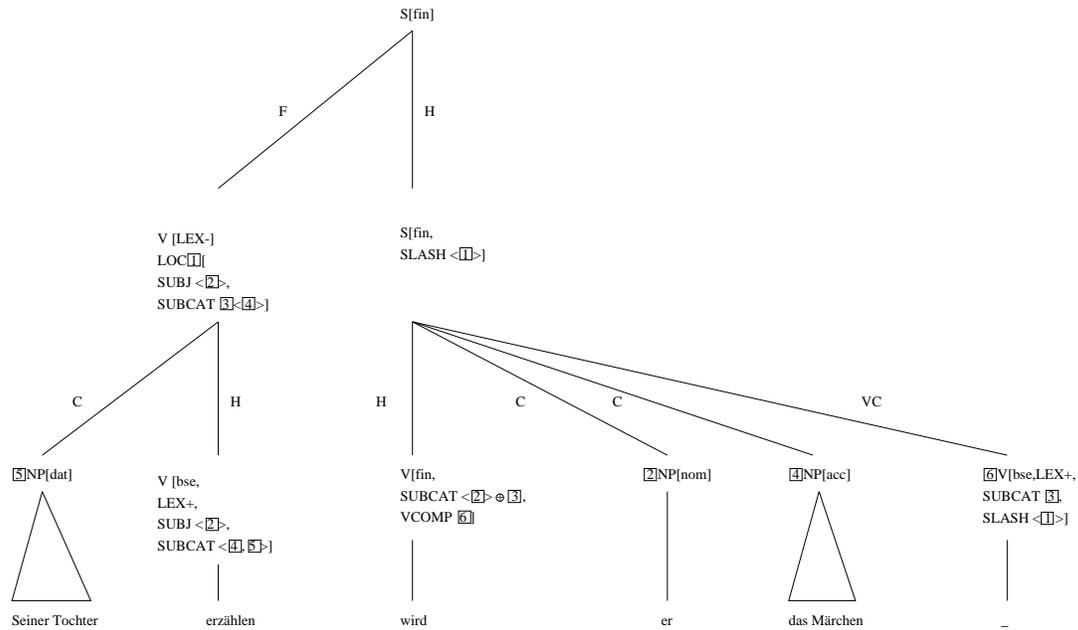}}}
\caption{\label{abb-seiner-tochter-erzaehlen} Analysis of {\em Seiner Tochter erz\"ahlen wird er das M\"archen.\/}}
\end{figure*}

\begin{figure*}[htbp]
\epsfxsize=0.9\textwidth
\centerline{\mbox{\epsffile{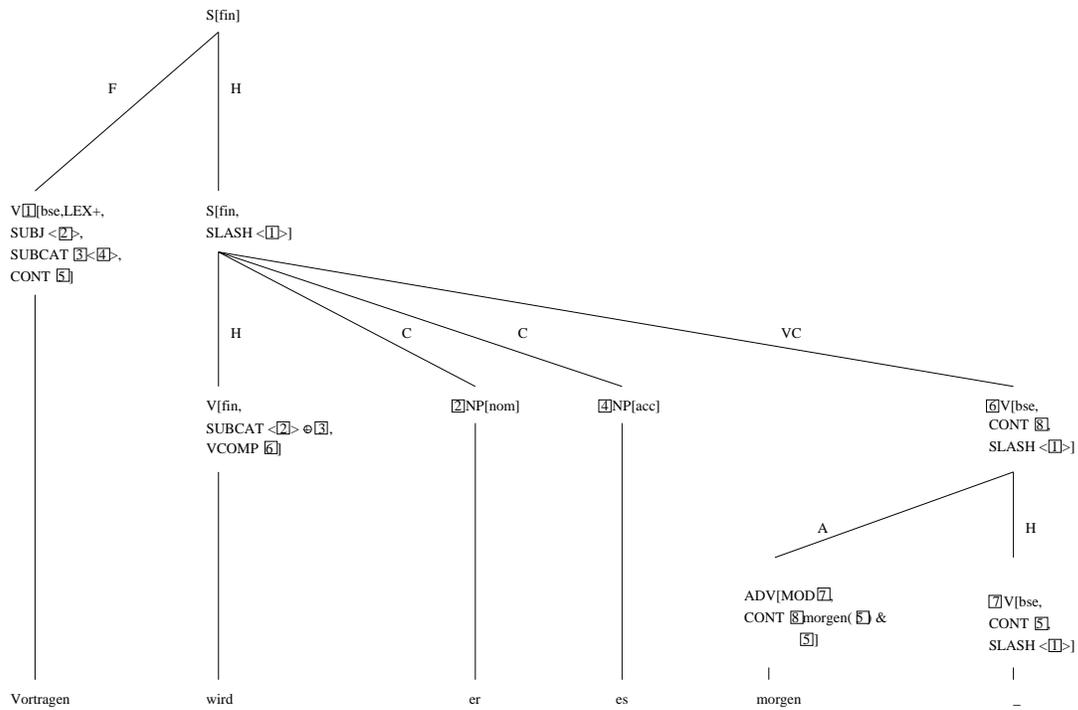}}}
\caption{\label{abb-vortragen-wird} Analysis of {\em Vortragen wird er es morgen.\/}}
\end{figure*}

Figures \ref{abb-seiner-tochter-erzaehlen} and \ref{abb-vortragen-wird} show the
analyses of the sentences in (\ex{1}).\footnote{
	In the original grammar, I use a binary branching schema for head-complement 
	and verb cluster structures. Adjuncts and complements are inserted into the domain
	of their head so that word order facts are accounted for.
	Due to space limitations, the figures show a tree for a flat head-complement structure.
} 
In the analyses of (\ex{1}a), a trace functions
as a verbal complement. In (\ex{1}b) a trace for a verb is modified by an adverb.
\eenumsentence{
\item Seiner Tochter erz\"ahlen wird er das M\"archen.
\item Vortragen wird er es morgen.
}


Sentences like (\ref{ex-muessen}a) are ruled out because {\em wird\/} selects a complement
in {\it bse\/}-form that has a {\sc vcomp} value {\it none\/}. As {\em erz\"ahlen\/} does not appear
in any {\sc comps} list it is not possible for the verb to count as an argument of the fronted
verbal complex that is saturated in the {\it Mittelfeld\/}. This is the case in Pollards
account. Hinrichs and Nakazawa have to block this case by stating type constraints on lists
of attracted arguments.
With a separate {\sc vcomp} feature this problem disappears.

\subsection{The Problem of Underspecified {\sc comps} Lists}
\label{licensing}

In this section, I will address a problem that seems to have gone unnoticed until now.
All analyses that involve argument attraction admit signs with underspecified {\sc comps} lists.
So in (\ref{bsp-erzaehlen-wird}), {\em wird\/} is combined with a trace or a lexical rule is
applied to it. The {\sc loc} value of the verbal complement is put into {\sc slash} and the arguments of the
verbal complement are attracted by the matrix verb. This list of arguments, however, is not
instantiated in the resulting sign. It remains variable until the {\sc slash} element becomes bound.
Therefore, the HPSG principles admit any kind of combination of totally unrelated signs.
Since the {\sc comps} list of the head is variable, any constituent is a possible complement.\footnote{
	The same problem exists for analysises that treat verb second as verb movement \cite{KW91,Netter92}.
}
As an HPSG theory is assumed to be a set of constraints that describe well formed descriptions
of linguistic objects, this is clearly not wanted. If a grammar contains phonologically empty
elements (traces, relativizers, and the like) the set of ill-formed signs will be infinite
because {\em wird \_$_i$\/} could be combined with arbitrarily many empty elements.\footnote{
        For a bottom-up parser, this would mean non-termination.
        }

It is clear that we want the matrix verb to behave in a very well defined way. It shall
attract exactly the arguments of the fronted verbal projection that were not saturated 
by this projection, i.e., the matrix verb shall perform the argument attraction that would
take place in base position, abstracting away from the value of {\sc lex}.
The desired effect can be reached if a rule schema is used for the introduction of nonlocal
dependencies. To introduce a nonlocal dependency for a verbal complex, this schema requires
an additional licensing condition to be met. The extracted element is licensed by an actually
existing verbal projection in the string. When a hearer of a sentence hears the words
that have to be combined with a trace or introduce the nonlocal dependency in another way,
he or she has already heard the phrase actually located in the {\it Vorfeld\/}. Therefore,
the information about the nonlocal dependency is present and
can be used to license the extracted element. The {\sc comps} list of the extracted element 
therefore is specified. The specified {\sc comps} are attracted by the matrix verb and the {\sc comps}
list of the matrix verb therefore does not contain any variables and our theory does
not admit signs that don't describe linguistic objects.

\newsavebox{\boxschema}
\sbox{\boxschema}{
\ms{ loc &    \ms{ cat$|$vcomp & none \\
                          } \\[3mm]
              nonloc & \ms{ inher$|$slash & \menge{ \ibox{1} } \\
                          } \\[3mm]
              lex & {\rm +} \\
            } 
}
\newsavebox{\boxschemab}
\sbox{\boxschemab}{
\ms[complement-slash-licencing-structure]{
           head-dtr  & \ms{ \ldots$|$vcomp$|$loc & \ibox{1} \\[1mm]
                            dom & \ibox{2} \\
                          } \\[3mm]
           vcomp-dtr & \ms{ \ldots \ms{ loc  \ibox{1}  \\[1mm]
                                        nl \ldots$|$sl \menge{}  \\
                                   $\!\!$} \\
                          $\!\!\!$} \\
         } 
}
\begin{samepage}
\begin{schema}[PVP-{\sc slash}-Introduction-Schema]
\label{pvp-trace-intro-schema-comp}
$
\\
\ms[phrasal-sign]{
synsem \usebox{\boxschema}\\[13mm]
dtrs   \usebox{\boxschemab}\\[15mm]
dom    \ibox{2} & \\
}
$
\end{schema}
\end{samepage}
Schema \ref{pvp-trace-intro-schema-comp} shows how this is implemented. A verbal complement of
a matrix verb is saturated. The {\sc vcomp} value of the resulting sign is {\it none\/}.
The {\sc loc} value of the saturated verbal complement is moved into {\sc slash}. This {\sc loc} value is licensed
by another verbal projection that meets the local requirements of the matrix verb but may be
positioned in the {\it Vorfeld\/}. As there are no constraints for daughters to be adjacent to
each other, there may be an arbitrary number of constituents between the licensing daughter
and the head daughter. The licensing daughter has licensing function only and is not
inserted into the domain of the resulting sign (\ibox{2}) at this point of combination.
However, an appropriate sign is inserted into the domain of its head when the nonlocal dependency
is bound.

\section{Alternatives}
\label{alternatives}
 
The drawback of the approaches of Pollard (To appear) and Nerbonne (1994)
are discussed in (Hinrichs and Nakazawa, 1994b). I will not repeat the arguments against these
approaches here. Instead, I will explain some of the problems of the Hinrichs and Nakazawa
approach.\footnote{
	Due to space limitations, I cannot give a detailed discussion of their approach here.
	The interested reader is referred to \cite{Mueller96-Eng-full-coling}.
}
 
Hinrichs and Nakazawa changed the value of {\sc slash}
into a set of signs rather than {\it local\/} objects.
The fronted phrase is a maximal projection with the missing constituents moved
to {\sc slash}. The fronted partial phrase is the filler for a nonlocal dependency 
which is introduced by their PVP-Topicalization Lexical Rule. As {\sc slash} elements are signs,
the lexical rule can refer to the {\sc slash} set of a {\sc slash} element and it is thus
possible to establish a relation between the {\sc comps} list of the auxiliary
and the {\sc slash} set of the fronted verbal projection. However, the assumption
that {\sc slash} contains signs rather than local objects is a change of the
basic HPSG formalism with far reaching consequences that is not really needed
and that has some side effects.
 
In the following, I discuss two problems for this approach.
Firstly, it is not possible to account for cases where a modifier in the {\it Mittelfeld\/}
modifies the fronted verbal projection without assuming an infinite lexicon because the only way
for a modifier to stay in the {\it Mittelfeld\/} while the modified constituent is fronted
is that the modifier is contained in the {\sc slash} set of the fronted constituent. It
therefore had to be a member of the {\sc comps} list.
An infinite lexicon is both
not very nice from a conceptual point of view and an implementational problem. Without
a complex control strategy (late evaluation) it is not possible to implement an infinite
lexicon. Another problem that was pointed out by Hinrichs and Nakazawa themselves is sentences like (\ex{1}).
\enumsentence{
\longex{5}{2}
        { * Gewu\ss{}t, & da\ss & Peter \_$_i$ & schl\"agt, & habe }{ ich & sie$_i$.}
        { ~~known       & that  & Peter        & hit        & have }{ I   & her     }
        {   `I knew that Peter hit her.'}
}
In (\ex{0}), {\em sie\/} is extracted from the complement sentence of {\em gewu\ss{}t\/} and than
inserted into the {\sc comps} list of {\em habe\/} and saturated in the {\it Mittelfeld\/}.
The same problem arises for other constructions involving nonlocal dependencies.\footnote{
        For an analysis of stranded prepositions in terms of nonlocal dependencies
        see \cite{Rentier94b} and \cite{Mueller95c-Eng}.
}
\eenumsentence{
\item \shortex{5}
	{[Da]$_i$ & hatte & Karl & [ \_$_i$ mit] & gerechnet.}
        {~this    & had   & Karl & ~with         & counted}
	{`Karl expected this.'}
\item * [ [ \_$_i$ mit] gerechnet ] hatte [da]$_i$ Karl.
}
\eenumsentence{
\item \shortex{4}
        {Bus$_i$ & will  & Karl [ \_$_i$ & fahren].}
        {bus     & wants & Karl          & drive}
        {`Karl wants to go by bus.'}
\item * [ \_$_i$ fahren] will Karl bus$_i$.
}

\section{Conclusion}

A very simple solution for the PVP problem was found. A minor change in the
feature geometry of signs was sufficient to cope with the spurious ambiguity problem
of Pollard's \shortcite{Pollard90} account. The account argued for in this paper can describe the fronting
phenomena without the assumption of an infinite lexicon. A solution for the problem
of underspecified {\sc comps\/} lists was found. This solution makes use of a 
schema to introduce the nonlocal dependency. An introduced nonlocal dependency is
licensed by an actually present element in the syntax analysis of a string.
At the point of combination, this element plays a licensing role
only and does not appear in the surface string of the build sign.
This is possible because two different levels of representation for combinatorial
and order information are used.

The analysis is part of an implemented fragment of German \cite{Babel-Eng}.

\bibliographystyle{acl}

\end{document}